\documentclass{article}

\usepackage{arxiv}

\usepackage[utf8]{inputenc} 
\usepackage[T1]{fontenc}    
\usepackage{hyperref}       
\usepackage{url}            
\usepackage{booktabs}       
\usepackage{amsfonts}       
\usepackage{nicefrac}       
\usepackage{microtype}      
\usepackage{lipsum}

\usepackage{times}
\usepackage{epsfig}
\usepackage{graphicx}
\usepackage{amsmath}
\usepackage{amssymb}
\usepackage{float}

\usepackage{times}
\usepackage{soul}
\usepackage{url}
\usepackage[small]{caption}
\usepackage{graphicx}
\usepackage{amsmath}
\usepackage{amsthm}
\usepackage{booktabs}
\usepackage{algorithm}
\usepackage{algorithmic}

\usepackage{color} 
\definecolor{lynn}{rgb}{0.8,0.5,0}
\usepackage[table,xcdraw]{xcolor} 
\usepackage{tabularx} 

\usepackage{adjustbox} 
\usepackage{multirow} 
\usepackage{subcaption}
\usepackage{authblk}

\title{Enabling Retrain-free Deep Neural Network Pruning using Surrogate Lagrangian Relaxation}

\author{Deniz Gurevin$^{1}$, Shanglin Zhou$^{2}$,  Lynn Pepin$^{2}$, Bingbing Li$^{2}$, Mikhail Bragin$^{1}$, Caiwen Ding$^{2}$, Fei Miao$^{2}$ \\
$^{1}$ Department of Electrical and Computer Engineering, University of Connecticut, USA\\
$^{2}$ Department of Computer Science and Engineering, University of Connecticut, USA\\
\texttt{\{deniz.gurevin, shanglin.zhou, lynn.pepin, bingbing.li, mikhail.bragin, caiwen.ding, fei.miao\}@uconn.edu}
}

\begin{document}

\maketitle

\begin{abstract}
Network pruning is a widely used technique to reduce computation cost and model size for deep neural networks. However, the typical three-stage pipeline, i.e., training, pruning and retraining (fine-tuning) significantly increases the overall training trails. In this paper, we develop a systematic weight-pruning optimization approach based on Surrogate  Lagrangian  relaxation (SLR), which is tailored to overcome difficulties caused by the discrete nature of the weight-pruning  problem  while  ensuring  fast  convergence. 
We further accelerate the convergence of the SLR by using quadratic penalties. Model parameters obtained by SLR during the training phase are much closer to their optimal values as compared to those obtained by other state-of-the-art methods.
We evaluate the proposed method on image classification tasks, i.e., ResNet-18 and ResNet-50 using ImageNet, and ResNet-18, ResNet-50 and VGG-16 using CIFAR-10, as well as object detection tasks, i.e., YOLOv3 and YOLOv3-tiny using  COCO  2014 and Ultra-Fast-Lane-Detection using TuSimple lane detection dataset.
Experimental results demonstrate that our SLR-based weight-pruning optimization approach achieves higher compression rate than state-of-the-arts under the same accuracy requirement. It also achieves a high model accuracy even at the hard-pruning stage without retraining (reduces the traditional three-stage pruning to two-stage). Given a limited budget of retraining epochs, our approach quickly recovers the model accuracy.


\end{abstract}


\section{Introduction} \label{introduction}


Deep neural network (DNN)-based statistical models are increasingly taxing of computational and storage resources, with costs proportional to the model size (i.e. the number of parameters in a model). This concern is especially pressing for embedded or IoT devices~\cite{krizhevsky2012imagenet,simonyan2014,DeepCompression}. By reducing model size, one reduces both storage costs and computation costs when evaluating a model. Various techniques exist for reducing model size while maintaining performance of the model, e.g. weight pruning, sparsity regularization, quantization, clustering. These techniques are collectively known as \emph{model compression} ~\cite{dai2017,yang2016,molchanov2017variational,guo2016dynamic,tung2017fine,luo2017thinet,DeepCompression,zhou2016less,DeepCompression,park2017weighted,frankle2018lottery,He_2018_ECCV,liu2018rethinking}.

 These works leverage the observation that training a compact model from scratch is more difficult and less performant than retraining a pruned model~\cite{li2016pruning,frankle2018lottery,liu2018rethinking}.  Therefore, a typical three-stage pipeline, i.e., training (large model), pruning and retraining (also called ''fine-tuning") is required. The pruning process is to set the redundant weights to zero and keep the important weights to best preserve the accuracy. The retraining process is necessary since the model accuracy dramatically drops after hard-pruning. However, this three-stage weight pruning technique significantly increases the overall training cost. For example, although the state-of-the-art weight pruning methods
 achieve very high compression rate while maintaining the prediction accuracy on many DNN architectures, the retraining process takes a longer time, e.g., 80 epochs for ResNet-18 on ImageNet, which is 70\% of the original training epochs using Alternate Direction Method of Multipliers (ADMM)~\cite{zhang2018systematic,ren2019admm}.

Given the pros and cons of current weight pruning-method, this paper aims to answer the following questions: 
\textbf{Is there an optimization method
that can achieve a high model accuracy even at the hard-pruning stage and can reduce retraining trails significantly?} \textbf{Given a limited budget of retraining epochs, is there an optimization method that can quickly recover the model accuracy (much faster than the state-of-the-art methods)?}
The major obstacle when answering these questions is the discrete nature of the model compression problems caused by ''cardinality" constraints, which ensure that a certain proportion of weights is pruned. 

In this paper, to address this difficulty, we develop a weight-pruning optimization approach based on recent Surrogate Lagrangian relaxation (SLR) \cite{Bragin2015SLR}, which overcomes all major convergence difficulties of standard Lagrangian relaxation. Lagrangian multipliers within SLR approach their optimal values much faster as compared to those within other methods.
We summarize our contributions/findings as:
\begin{itemize}
\item Novel SLR-based approach tailored to overcome difficulties caused by the discrete nature of the weight-pruning problem while ensuring fast convergence. 
\item We further accelerate the convergence of the SLR by using quadratic penalties. The method possesses nice convergence properties inherited from fast accelerated reduction of constraint violation
due to quadratic penalties and from the guaranteed convergence thereby leading to unparalleled performance as compared to other methods. Therefore, model parameters obtained by SLR
are much closer to their optimal values as compared to those obtained by other state-of-the-art methods.
\item 
Existing coordination-based weight pruning methods do not converge when solving non-convex problems while SLR does. 
Other coordination methods (e.g., ADMM) are not designed to handle discrete variables and other types of non-convexities.

\item Our SLR-based weight-pruning optimization approach achieves a high model accuracy even at the hard-pruning stage and given a limited budget of retraining epochs, it quickly recovers the model accuracy.


\end{itemize}

\section{Related Works on Weight Pruning}

Since lots of researchers have investigated that some portion of weights in neural networks are redundant, weight pruning is proposed to remove these less important coefficient values and it achieves model compression with similar performance compared to uncompressed one. 
Early work~\cite{han2015learning} proposed an iterative
and static magnitude-based weight pruning to explore the redundancy. Later, the compression rate has been further improved by integrating the ADMM~\cite{ADMMBoyd2011},
an optimization algorithm that breaks optimization problems into smaller subproblems, where dynamic penalty is applied on all targeted weights each of which is then solved iteratively and more easily~\cite{zhang2018systematic,niu2020patdnn}. 
\cite{louizos2018learning} proposed a framework for $L_0$ norm regularization for neural networks, which aim to prune the network during training by choosing weights setting them to exactly zero. 
Lottery ticket hypothesis was proposed by ~\cite{frankle2018lottery}, which observes that a subnetwork of randomly-initialized network can replace the original network with the same performance.

\section{Weight Pruning using SLR}
Consider a DNN with $N$ layers indexed by $n \in 1,...,N$, where the weights 
at layer $n$ are denoted by ${{\bf W}}_{n}$.
The objective is to minimize the loss function
subject to constraints on the cardinality of weights within each layer $n$ (the number of nonzero weights
should be less than or equal to the predefined number $l_n$): 
$ \underset{ {\bf{W}}_{n}}{\text{min}}
\ \ \  f \big( {\bf{W}}_{n}\big) +
     \sum_{n=1}^{N} g_{n}({\bf{W}}_{n})$, 
where the first term represents the nonlinear smooth loss function and the other represents the non-differentiable ``cardinality" penalty term~\cite{zhang2018systematic}. $g_{n}(\cdot)$ is the indicator function:
\begin{eqnarray*}g_{n}({\bf{W}}_{n})=
\begin{cases}
 0 & \text {if } \mathrm{card}({\bf{W}}_{n})\le l_{n}, \; n = 1, \ldots, N, \\
 +\infty & \text {otherwise. } 
\end{cases}
\end{eqnarray*}

 In its entirety, the problem cannot be solved either analytically or by using the stochastic gradient descent.  To enable the decomposition into smaller manageable subproblems, duplicate variables are introduced and the problem is equivalently rewritten as:
$\underset{ {\bf{W}}_{n}}{\text{min}}
 f \big( {\bf{W}}_{n} \big)+\sum_{n=1}^{N} g_{n}({\bf{Z}}_{n}), \label{losscardinality}
 \text{subject to}~{\bf{W}}_{n}={\bf{Z}}_{n}, \; n = 1, \ldots, N.$
To solve the problem,
constraints
are first relaxed by introducing Lagrangian multipliers and their violations are penalized by using quadratic penalties.  The resulting \textit{Augmented} Lagrangian function ~\cite{ADMMBoyd2011,zhang2018systematic} of the above optimization problem is this given by
\begin{equation}
\begin{aligned}
 L_{\rho} &   \big( {\bf{W}}_{n}, {\bf{Z}}_{n} , {\bf{\Lambda}}_{n}  \big) = 
   f \big( {\bf{W}}_{n} \big) \label{relaxedproblem}
 + \sum_{n=1}^{N} g_{n}({\bf{Z}}_{n}) +\sum_{n=1}^{N} \mathrm{tr} [{\bf{\Lambda}}_{n}^T({\bf{W}}_{n}-{\bf{Z}}_{n}) ]
 + \sum_{n=1}^{N} \frac{\rho}{2} \| {\bf{W}}_{n}-{\bf{Z}}_{n} \|_{F}^{2},
\end{aligned}
\end{equation}
where ${\bf{\Lambda}}_{n}$ is a matrix of Lagrangian multipliers (dual variables) corresponding to constraints ${\bf{W}}_{n}={\bf{Z}}_{n}$, and has the same dimension as ${\bf{W}}_{n}$. The positive scalar $\rho$ is the penalty coefficient, $\mathrm{tr}(\cdot)$ denotes the trace, and $ \| \cdot \|_{F}^{2}$ denotes the Frobenius norm.

In the following, motivated by decomposibility enabled by SLR~\cite{Bragin2015SLR}, which overcame all major difficulties of standard Lagrangian Relaxation, with much alleviated zigzagging and guaranteed convergence, the relaxed problem will be decomposed into two manageable subproblems, and the subproblems will then be efficiently coordinated by Lagrangian multipliers.


\noindent \textbf{Step 1: Solve ``Loss Function" Subproblem for ${\bf{W}}_{n}$ by using Stochastic Gradient Decent.} At iteration $k$, for given values of multipliers ${\bf{\Lambda}}_{n}^k$, the first ''loss function" subproblem is to minimize the Lagrangian function, while keeping ${\bf{Z}}_{n}$ at previously obtained values ${\bf{Z}}_{n}^{k-1}$ as

\begin{equation}
\begin{aligned}
 \min_{ {\bf{W}}_{n}} L_{\rho} &   \big( {\bf{W}}_{n}, {\bf{Z}}_{n}^{k-1} , {\bf{\Lambda}}_{n} \big). \label{lossfunctionsubproblem}
\end{aligned}
\end{equation}

\noindent Since the regularizer is a differentiable quadratic norm, and the loss function is differentiable, the subproblem can be solved by stochastic gradient descent (SGD)~\cite{bottou2010large}.  To ensure that multipliers updating directions are ``proper," the following ``surrogate" optimality condition needs to be satisfied following \cite[p. 179, eq. (12)]{Bragin2015SLR}:



  \begin{equation}
\label{SOC1}
L_{\rho} \big({\bf{W}}_{n}^{k}, {\bf{Z}}_{n}^{k-1}, {\bf{\Lambda}}_{n}^{k} \big) < 
L_{\rho} \big({\bf{W}}_{n}^{k-1}, {\bf{Z}}_{n}^{k-1}, {\bf{\Lambda}}_{n}^{k} \big). 
\end{equation}
 
If \eqref{SOC1} is satisfied, multipliers are updated following \cite[p. 179, eq. (15)]{Bragin2015SLR} as: 
\begin{equation} 
\begin{aligned}
{\bf{\Lambda}'}_{n}^{k+1}: = &{\bf{\Lambda}}_{n}^{k}+ s'^k({\bf{W}}_{n}^{k}-{\bf{Z}}_{n}^{k-1}).\label{multiplierupdate1}
\end{aligned}
\end{equation}
\noindent where stepsizes are updated as \cite[p. 180, eq. (20)]{Bragin2015SLR} as
\begin{equation}
\begin{aligned}
s'^{k} =& \alpha^{k} \frac{s^{k-1}||{\bf{W}}^{k-1}-{\bf{Z}}^{k-1}||}{||{\bf{W}}^{k}-{\bf{Z}}^{k-1}||}. \label{step1}
\end{aligned}
\end{equation}

\noindent \textbf{Step 2: Solve ``Cardinality" Subproblem for ${\bf{Z}}_{n}$ through Pruning by using Projections onto Discrete Subspace.} The second ``cardinality" subproblem is solved with respect to ${\bf{Z}}_{n}$ while fixing other variables at values ${\bf{W}}_{n}^k$ as  
\begin{equation}
\begin{aligned}
 \min_{ {\bf{Z}}_{n}} L_{\rho} &   \big( {\bf{W}}_{n}^k, {\bf{Z}}_{n}, {\bf{\Lambda}'}_{n}^{k+1} \big). \label{cardinalitysubproblem}
\end{aligned}
\end{equation}

\noindent Since $g_{n}(\cdot)$ is the indicator function, the globally optimal solution of this problem can be explicitly derived as \cite{ADMMBoyd2011}:
\begin{equation}
\label{5}
  {\bf{Z}}_{n}^{k} = {{\bf{\Pi}}_{{\bf{S}}_{n}}}\big({\bf{W}}_{n}^{k}+\frac{{\bf{\Lambda}'}_{n}^{k+1}}{\rho}\big),
\end{equation}
where ${{\bf{\Pi}}_{{\bf{S}}_{n}}(\cdot)}$ denotes the Euclidean projection onto the set ${\bf{S}}_{n}= \{{\bf{W}}_{n}\mid \mathrm{card}({\bf{W}}_{n})\le l_{n}  \}, n=1,\dots,N$.


 To ensure that multipliers updating directions are ``proper," the following ``surrogate" optimality condition needs to be satisfied:
  \begin{equation}
\label{SOC}
 L_{\rho} \big({\bf{W}}_{n}^{k} , {\bf{Z}}_{n}^{k}, {\bf{\Lambda}'}_{n}^{k+1}  \big) < 
 L_{\rho} \big({\bf{W}}_{n}^{k} , {\bf{Z}}_{n}^{k-1} , {\bf{\Lambda}'}_{n}^{k+1} \big)
\end{equation}
 
\noindent Once \eqref{SOC} is satisfied,\footnote{If condition \eqref{SOC} is not satisfied, the subproblems \eqref{lossfunctionsubproblem} and \eqref{cardinalitysubproblem} are solved again by using the latest available values for  ${\bf{W}}_{n} $ and  ${\bf{Z}}_{n}$.} multipliers are updated as:

\begin{equation} 
\begin{aligned}
{\bf{\Lambda}}_{n}^{k+1}: = &{\bf{\Lambda}'}_{n}^{k+1}+ s^k({\bf{W}}_{n}^{k}-{\bf{Z}}_{n}^{k}),\label{multiplierupdate2}
\end{aligned}
\end{equation}


\noindent where stepsizes are updated as 
\begin{equation}
\begin{aligned}
s^{k} =& \alpha^{k} \frac{s'^{k}||{\bf{W}}^{k-1}-{\bf{Z}}^{k-1}||}{||{\bf{W}}^{k}-{\bf{Z}}^{k}||}, \label{step1}
\end{aligned}
\end{equation}
where stepsize-setting parameters \cite[p. 188, eq. (67)]{Bragin2015SLR} are:
\begin{equation}
\begin{aligned}
\alpha^{k} = & 1 - \frac{1}{M \times k^{(1-\frac{1}{k^r})}}, M > 1, 0 < r < 1.
\end{aligned} \label{step2}
\end{equation}

The algorithm of the new method is presented below:

\begin{algorithm}[H] 
\caption{Surrogate Lagrangian Relaxation}
\begin{algorithmic}[1] 
\STATE Initialize ${\bf{W}}_{n}^{0} , {\bf{Z}}_{n}^{0} , {\bf{\Lambda}}_{n}^{0} $ and $s^0$ \\
\WHILE{Stopping criteria are not satisfied}
     \STATE \textbf 1 solve subproblem \eqref{lossfunctionsubproblem}, \;
  
  \IF{surrogate optimality condition \eqref{SOC1} is satisfied }
  \STATE keep ${\bf{W}}_{n}^{k} , {\bf{Z}}_{n}^{k} $, and update multipliers ${\bf{\Lambda}}_{n}^{k} $ per \eqref{multiplierupdate1} ,
   
   \ELSE
   \STATE  keep ${\bf{W}}_{n}^{k} , {\bf{Z}}_{n}^{k} $, do not update multipliers ${\bf{\Lambda}}_{n}^{k} $,

   \ENDIF

         \STATE \textbf 2 solve subproblem \eqref{cardinalitysubproblem}, \; 
    \IF{surrogate optimality condition \eqref{SOC} is satisfied }
  \STATE keep ${\bf{W}}_{n}^{k} ,  {\bf{Z}}_{n}^{k} $, and update multipliers ${\bf{\Lambda}}_{n}^{k} $ per \eqref{multiplierupdate2} ,

   \ELSE
   \STATE  keep ${\bf{W}}_{n}^{k} ,  {\bf{Z}}_{n}^{k} $, do not update multipliers ${\bf{\Lambda}}_{n}^{k} $, 
   \ENDIF

   \ENDWHILE
\end{algorithmic}
\end{algorithm}

\noindent The theoretical results are summarized below:

\noindent \textbf{Theorem. Sufficient Condition for Convergence of the Method:} Assuming for any integer number $\kappa$ there exists $k > \kappa$ such that surrogate optimality conditions \eqref{SOC1} and \eqref{SOC} are satisfied, then  under the stepsizing conditions \eqref{step1}-\eqref{step2}, the Lagrangian multipliers converge to their optimal values $\mathbf{\Lambda}_n^{*}$ that maximize the following dual function:  

  \begin{equation}
\label{qual}
q({\bf{\Lambda}}) \equiv \min_{{\bf{W}}_{n} , {\bf{Z}}_{n}} L_{\rho} \big({\bf{W}}_{n} , {\bf{Z}}_{n}, {\mathbf{\Lambda}}_{n}  \big).
\end{equation}


\begin{proof}
The proof will be based on that of \cite{Bragin2015SLR}.  The major difference in the original SLR method \cite{Bragin2015SLR} and the SLR method of this paper is the presence of quadratic terms within the Lagrangian function \eqref{relaxedproblem}.  

It should be noted that the weight pruning problem can be equivalently rewritten in a generic form as:

\begin{equation}
\begin{aligned}\min_{\bf{X}} \bf{F}(\bf{X}), 
\textit{s.t. }  \bf{G}(\bf{X}) = 0. \label{GeneralEquation}
 \end{aligned}
 \end{equation}
 where $\bf{X}$ collectively denotes the decision variables $\{{\bf{W}}_{n}, {\bf{Z}}_{n}\}$ and 
\begin{equation}
\begin{aligned}
& {\bf{F(X)}} \equiv 
   f \big( {\bf{W}}_{n} \big) 
 + \sum_{n=1}^{N} g_{n}({\bf{Z}}_{n}) + \sum_{n=1}^{N} \frac{\rho}{2} \| {\bf{W}}_{n}-{\bf{Z}}_{n} \|_{F}^{2},
 \\
& {\bf{G(X)}} \equiv {\bf{W}}_{n} - {\bf{Z}}_{n},  n = 1, \ldots, N. 
 \end{aligned}
 \end{equation}
 
 The feasible set of \eqref{GeneralEquation} is equivalent to that of the original model compression problem.
 Feasibility requires that ${\bf{W}}_{n} = {\bf{Z}}_{n}$, which makes the term $\frac{\rho}{2} \| {\bf{W}}_{n}-{\bf{Z}}_{n} \|_{F}^{2}$ within \eqref{GeneralEquation} disappear. Therefore, the Lagrangian function that corresponds to \eqref{GeneralEquation} is the \textit{Augmented} Lagrangian function \eqref{relaxedproblem} to the original model compression problem. Furthermore, the surrogate optimality conditions \eqref{SOC1} and \eqref{SOC} are the surrogate optimality conditions that correspond to the Lagrangian function $\bf{F(\bf{X})} + \bf{\Lambda} \bf{G(\bf{X})}$ that corresponds to the problem \eqref{GeneralEquation}.  Therefore, since within the original SLR \cite[Prop. 2.7, p. 188]{Bragin2015SLR} convergence was proved under conditions on stepsizes \eqref{step1}-\eqref{step2} and the satisfaction of surrogate optimality conditions, both of which are satisfied here, multipliers converge to their optimal values for the model compression under consideration as well.  
\end{proof}

This is a major result in machine learning since this is the first application of the SLR method to guarantee theoretical convergence of the model compression problem while handling discrete variables as well as the quadratic terms. In fact, owing to quadratic terms, the method inherits nice convergence properties similar to those of Augmented Lagrangian Relaxation (ALR) (fast reduction of constraint violations) and to those of SLR (fast and guaranteed convergence without much of zigzagging of multipliers and the need of the so-called ``optimal dual value" \cite{Bragin2015SLR}) thereby leading to unparalleled performance as compared to other methods. 

The SLR method enjoys the benefits of efficient subproblem solution coordination with guaranteed convergence made possible by stepsizes \eqref{step1}-\eqref{step2} approaching zero (without this requirement, multipliers \eqref{multiplierupdate2} would not exhibit convergence), and by the satisfaction of surrogate surrogate optimality conditions \eqref{SOC1} and \eqref{SOC} ensuring that multipliers are updated along ``good" directions. It will be verified empirically by the ``Evaluation" section that there always exists iteration $\kappa$ after which the sufficient conditions are satisfied thereby ensuring that the multipliers approach their optimal values during the entire iterative process.

The SLR method also benefits from the independent and systematic adjustment of two hyper-parameters: penalty coefficient $\rho$ and the stepsize $s^k$. In contrast, other coordination methods are not designed to handle discrete variables and other types of non-convexities.  For example, ADMM do not converge when solving non-convex problems \cite[p.~73]{ADMMBoyd2011} because stepsizes $\rho$ within the method does not converge to zero. Lowering stepsizes to zero within ADMM would lead to decrease of the penalty coefficient, thereby leading to slower convergence.

\section{Evaluation} \label{evaluation}


\noindent\textbf{Evaluation Metrics.}
We start by pruning the pretrained models through SLR training. 
Afterwards, we perform hard-pruning on the model, completing the compression phase. We report the overall compression rate (or the percentage of remaining weights) and prediction accuracy.

\noindent\textbf{Experimental Setup.}\label{image_setup}
\noindent All of the baseline models we use and our code in image classification tasks are implemented with PyTorch 1.6.0 and Python 3.6. For our experiments on COCO 2014 dataset, we used Torch v1.6.0, pycocotools v2.0 packages. For our experiments on TuSimple lane detection benchmark dataset\footnote{\url{https://github.com/TuSimple/tusimple-benchmark}}, we used Python 3.7 with Torch v1.6.0, and SpConv v1.2 package.  We conducted our experiments on Ubuntu 18.04 and using Nvidia Quadro RTX 6000 GPU with 24 GB GPU memory. We used 4 GPU nodes to train our models on the ImageNet dataset.

\subsection{Evaluation on Image Classification Tasks} \label{classification}

\noindent\textbf{Models and Datasets.} We compare our SLR method against ADMM. We tested our SLR method as well as the ADMM, by using 3 DNN models (ResNet-18, ResNet-50 \cite{he2016deep} and VGG-16 \cite{simonyan2014}) on CIFAR-10 and 2 DNN models (ResNet-18 and ResNet-50) on ImageNet ILSVRC 2012 benchmark. We use the pretrained ResNet models on ImageNet from Torchvision's ``models" subpackage. The accuracy of the pretrained baseline models we used are listed in Table~\ref{table:addmvsslr_classification}.

\noindent\textbf{Training Settings.} In all experiments we used $\rho = 0.1$. In CIFAR-10 experiments, we used a learning rate of $0.01$, batch size of 128 and ADAM optimizer during training. On ImageNet, we used a learning rate of $10^{-4}$, batch size of 256 and SGD optimizer. For a fair comparison of SLR and ADMM methods, we used the same number of training epochs and sparsity configuration for both methods in the experiments.\\

\noindent\textbf{Evaluation of SLR Performance.} Table~\ref{table:addmvsslr_classification} shows our comparison of SLR and ADMM on CIFAR-10 and ImageNet benchmark. Here, SLR parameters are set as  $M = 300$, $r = 0.1$ and $s_0 = 10^{-2}$. After SLR and ADMM training, final hardpruning is performed and the hardpruning accuracy is reported without any additional retraining, given a limited budget of training epochs. According to our results, SLR outperforms ADMM method in terms of accuracy under the same compression rate. With higher compression rates, CIFAR-10 results show higher gap in accuracy between SLR and ADMM. \\



\begin{table}[htb!]
\caption{Comparison of SLR and ADMM on CIFAR-10 and ImageNet datasets. ImageNet results are based on Top-5 accuracy.}
\label{table:addmvsslr_classification}
\centering
\begin{adjustbox}{width=0.6\textwidth}
\begin{tabular}{cccccc}
\hline
                           & \textbf{\begin{tabular}[c]{@{}c@{}}Baseline\\ Acc. (\%)\end{tabular}} & \textbf{Epochs}                           & \textbf{Method} & \textbf{\begin{tabular}[c]{@{}c@{}}Pruning \\ Acc. (\%)\end{tabular}} & \textbf{\begin{tabular}[c]{@{}c@{}}Compression \\ Rate\end{tabular}} \\ \hline
\textbf{CIFAR-10}          &                                                                       &                                           &                 &                                                                       &                                                                      \\ \hline
\multirow{2}{*}{ResNet-18} & \multirow{2}{*}{93.33}                                                & \multicolumn{1}{c|}{\multirow{2}{*}{50}}  & ADMM            & 72.84                                                                 & \multirow{2}{*}{8.71$\times$}                                        \\
                           &                                                                       & \multicolumn{1}{c|}{}                     & SLR             & \textit{89.93}                                                        &                                                                      \\ \hline
\multirow{2}{*}{ResNet-50} & \multirow{2}{*}{93.86}                                                & \multicolumn{1}{c|}{\multirow{2}{*}{50}}  & ADMM            & 78.63                                                                 & \multirow{2}{*}{6.57$\times$}                                        \\
                           &                                                                       & \multicolumn{1}{c|}{}                     & SLR             & \textit{88.91}                                                        &                                                                      \\ \hline
\multirow{2}{*}{VGG-16}    & \multirow{2}{*}{93.27}                                                & \multicolumn{1}{c|}{\multirow{2}{*}{110}} & ADMM            & 69.05                                                                 & \multirow{2}{*}{12$\times$}                                          \\
                           &                                                                       & \multicolumn{1}{c|}{}                     & SLR             & \textit{87.31}                                                        &                                                                      \\ \hline
\textbf{ImageNet}          &                                                                       &                                           &                 &                                                                       &                                                                      \\ \hline
\multirow{2}{*}{ResNet-18} & \multirow{2}{*}{89.07}                                                & \multicolumn{1}{c|}{\multirow{2}{*}{40}}  & ADMM            & 80.69                                                                 & \multirow{2}{*}{6.5$\times$}                                         \\
                           &                                                                       & \multicolumn{1}{c|}{}                     & SLR             & \textit{82.70}                                                         &                                                                      \\ \hline
\multirow{2}{*}{ResNet-50} & \multirow{2}{*}{92.87}                                                & \multicolumn{1}{c|}{\multirow{2}{*}{30}}  & ADMM            & 85.10                                                                  & \multirow{2}{*}{3.89$\times$}                                        \\
                           &                                                                       & \multicolumn{1}{c|}{}                     & SLR             & \textit{87.50}                                                         &                                                                      \\ \hline
\end{tabular}

\end{adjustbox}
\end{table}

\begin{figure}[htb!]
  \centering
  \subcaptionbox{ResNet-18. \label{fig:resnet18_cifar}}{\includegraphics[width=1.9in]{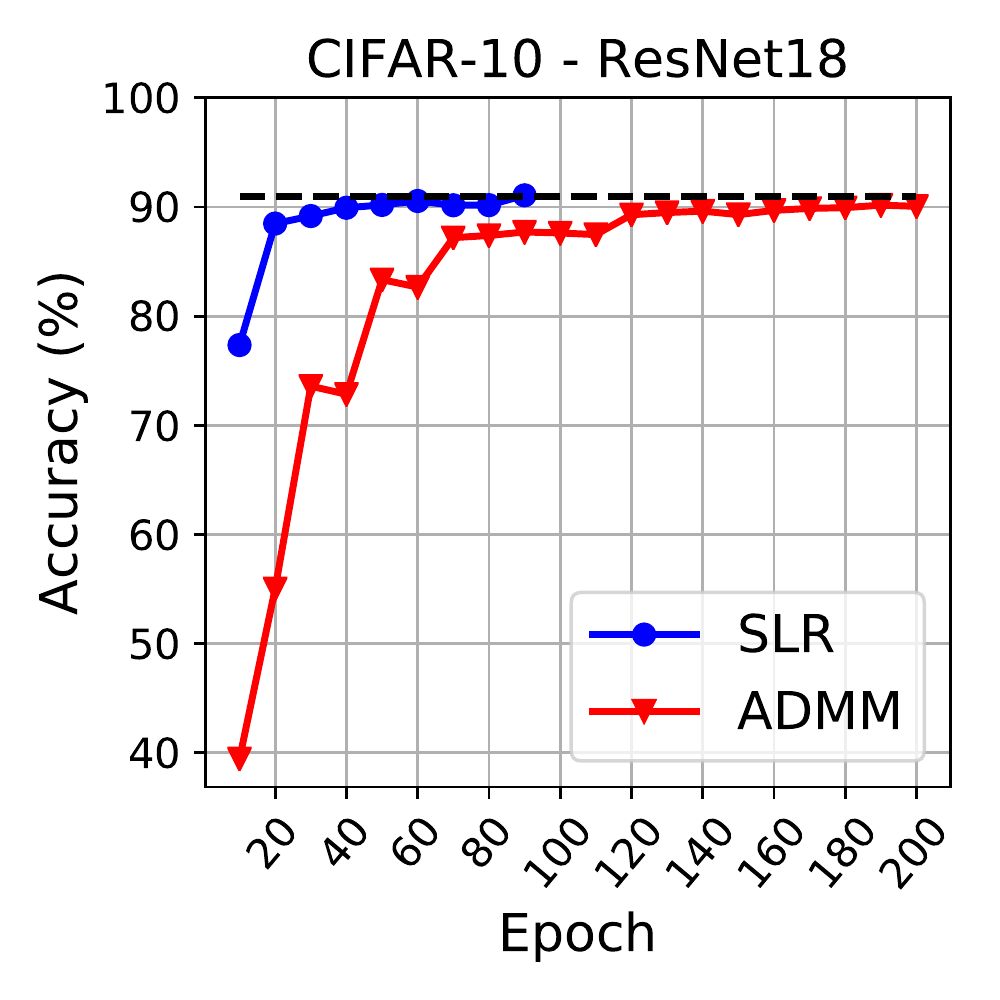}}\hspace{0em}%
  \subcaptionbox{ResNet-50. \label{fig:resnet50_cifar}}{\includegraphics[width=1.9in]{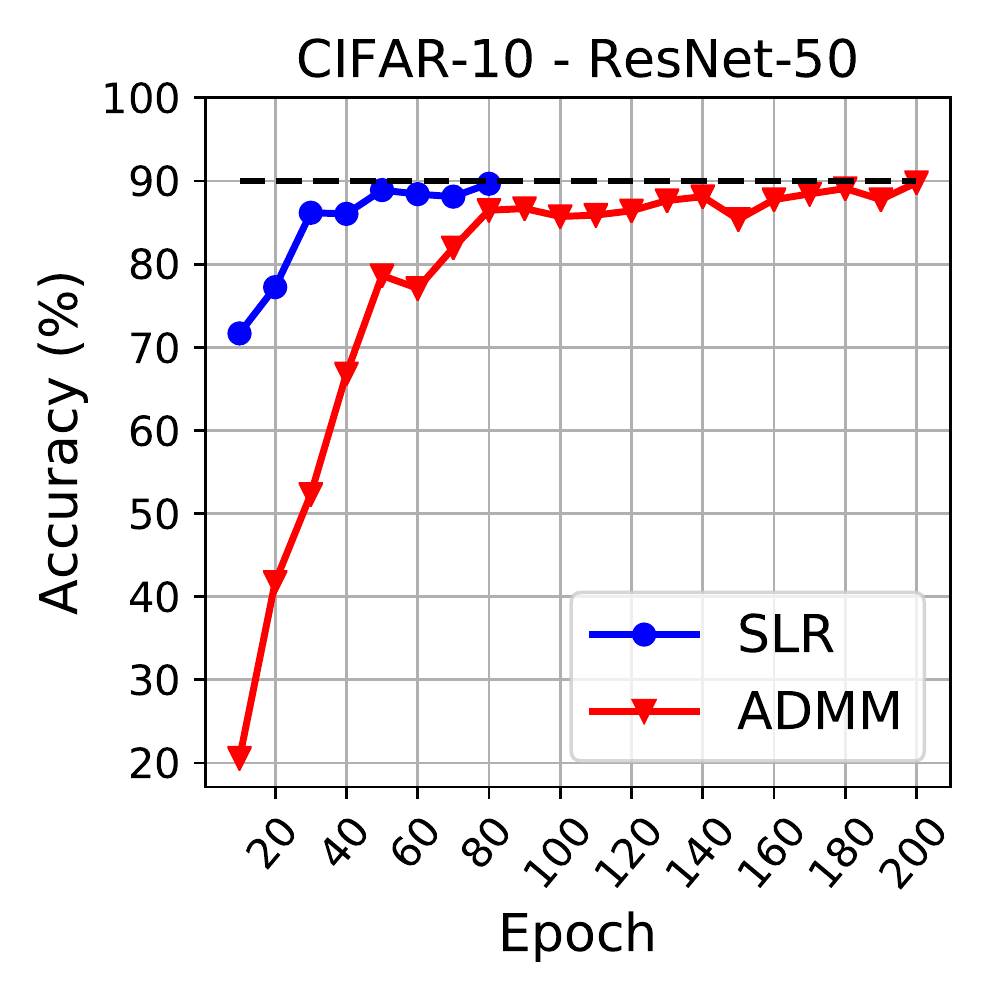}}\hspace{0em}%
  \subcaptionbox{VGG-16.\label{fig:vgg_cifar}}{\includegraphics[width=1.9in]{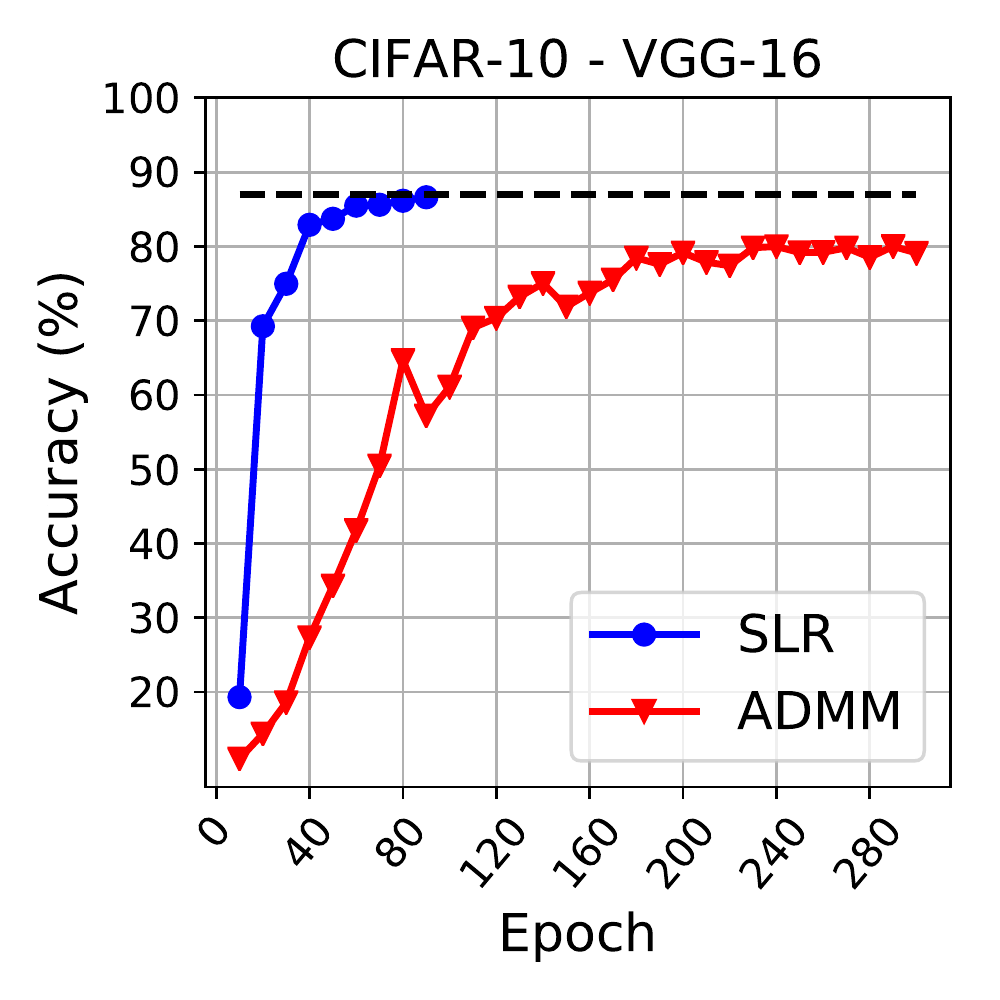}}

  \caption{Hardpruning accuracy after SLR vs. ADMM training on CIFAR-10. Accuracy is reported every 10 epochs and training is stopped when desired accuracy is reached. }
\label{fig:hardprune_cifar10}
\end{figure}

Figure \ref{fig:hardprune_cifar10} shows the hardpruning accuracy during SLR vs. ADMM on CIFAR-10 and ImageNet, corresponding to Table \ref{table:addmvsslr_classification}. During training, hardpruning accuracy is checked periodically. If the hardpruning accuracy meets the accuracy criteria, the training is stopped. As seen in the figures, SLR quickly converges and reaches the desired accuracy, almost 3$\times$ faster than ADMM on CIFAR-10. Moreover, in Figure \ref{fig:vgg_cifar}, ADMM is still below the desired accuracy even after 300 epochs of training on VGG-16, while SLR completes training in 80 epochs.\\

\begin{figure}[H]
  \centering
  \subcaptionbox{ResNet-18. \label{fig3:a}}{\includegraphics[width=2.6in]{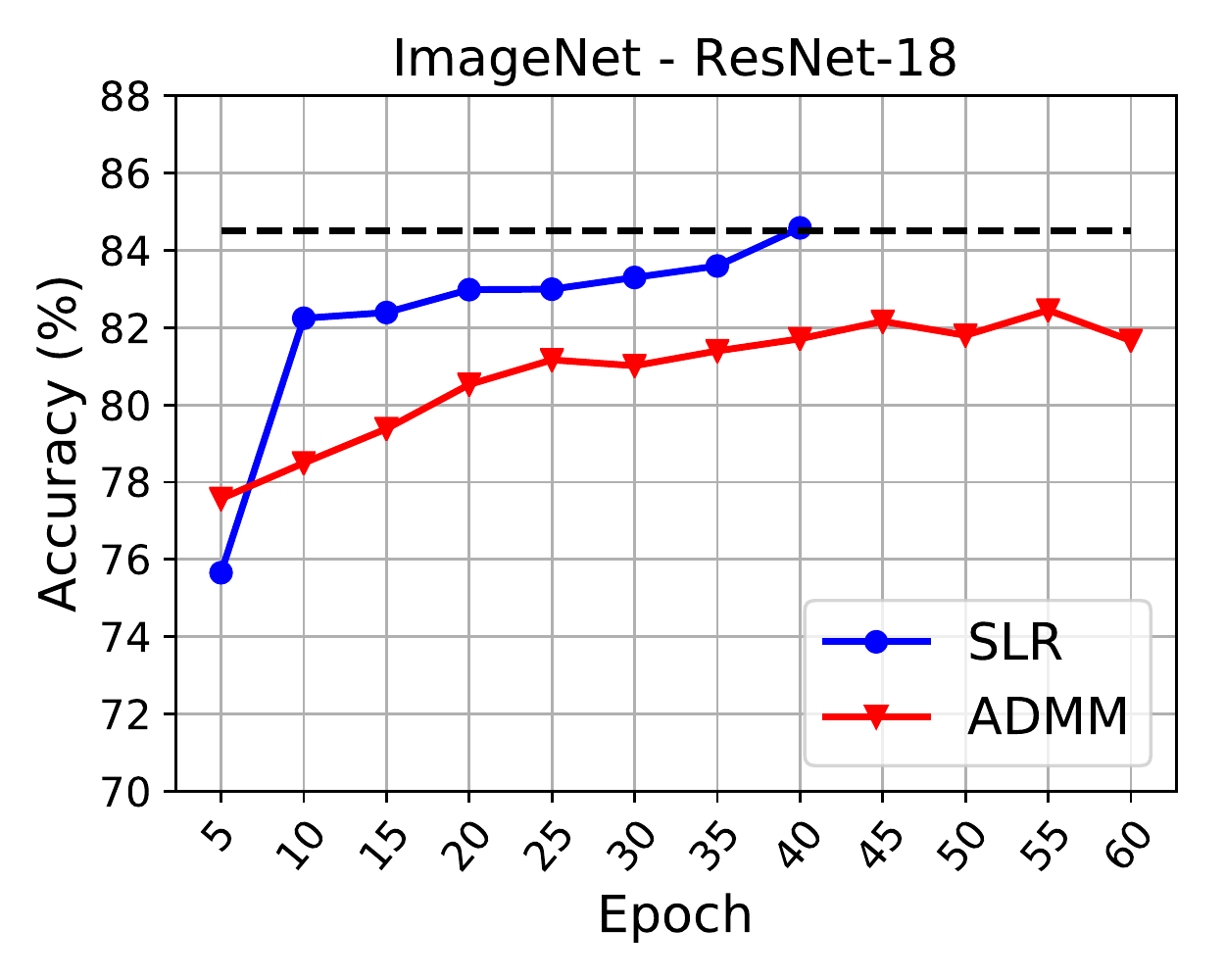}}\hspace{0em}%
  \subcaptionbox{ResNet-50. \label{fig3:b}}{\includegraphics[width=2.6in]{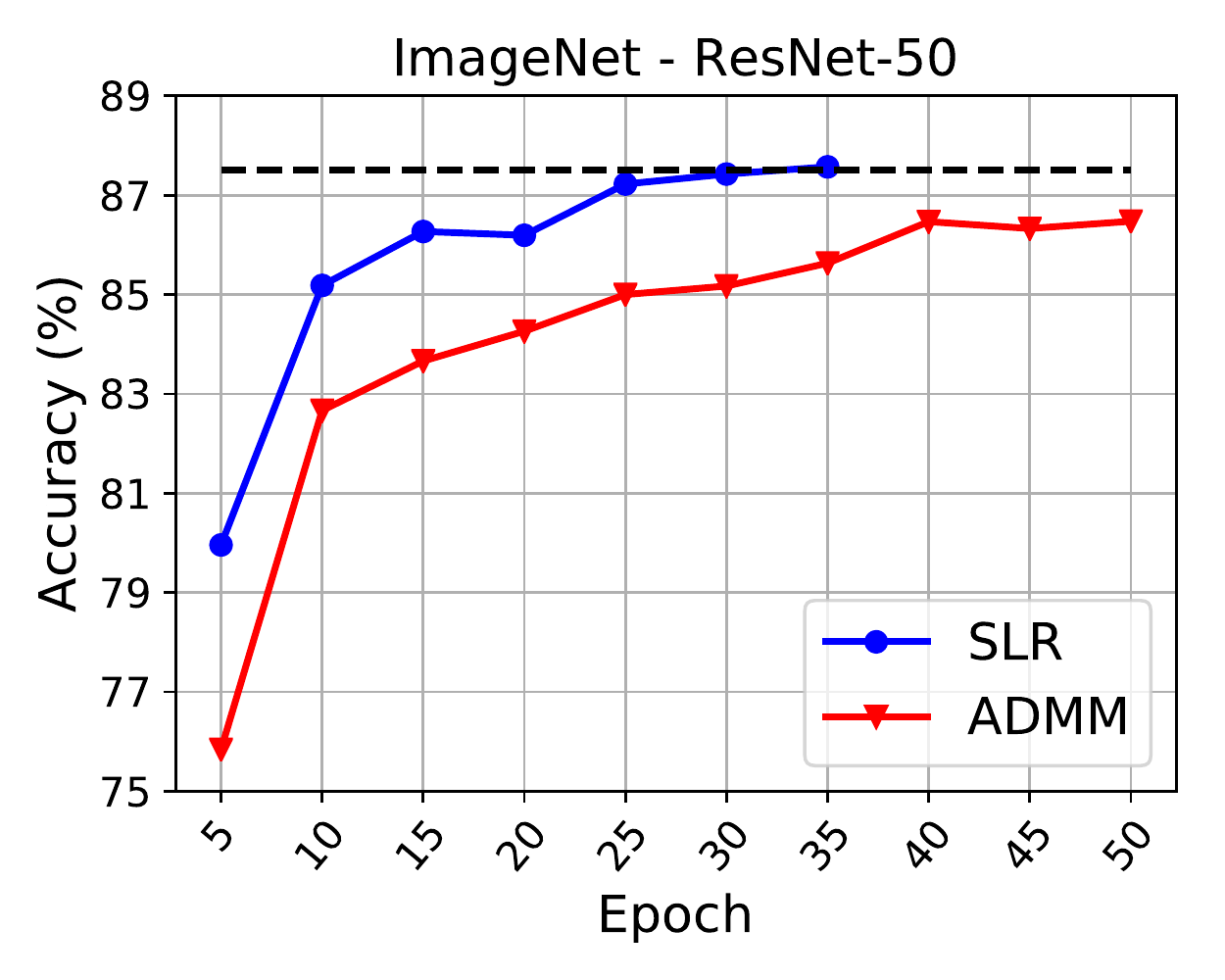}}

  \caption{Hardpruning accuracy after SLR vs. ADMM training on ImageNet. Accuracy is reported every 5 epochs and training is stopped when desired accuracy is reached.}
\label{fig:hardprune_imagenet}
\end{figure}

\noindent Similarly, as can be seen in Figure \ref{fig:hardprune_imagenet}, ADMM cannot output the desired accuracy after 60 and 50 epochs of training on ImageNet, while SLR reaches the threshold quickly.

\begin{table}[htb!]
\caption{SLR performance comparison with VGG-16, ResNet-18 and ResNet-50 on CIFAR-10 benchmark.}
\label{table:slr_comp}
\centering
\begin{adjustbox}{width=0.6\columnwidth}
\begin{tabular}{llcc}
\hline
\textbf{Model}             & \textbf{Method}         & \textbf{Accuracy} & \textbf{\begin{tabular}[c]{@{}c@{}}Weights\\ Remaining\end{tabular}} \\ \hline
\multirow{7}{*}{VGG-16}    & \textbf{SLR}            & 91.2\%            & \multirow{3}{*}{10\%}                                                \\
                           & AMC \cite{He_2018_ECCV}                     & 91.0\%            &                                                                      \\
                           & L0 \cite{louizos2018learning}                      & 80.0\%            &                                                                      \\ \cline{2-4} 
                           & \textbf{SLR}            & 93.1\%            & \multirow{2}{*}{40\%}                                                \\
                           & One-shot pruning\cite{liu2018rethinking}        & 92.4\%            &                                                                      \\ \cline{2-4} 
                           & \textbf{SLR}            & 93.2\%            & \multirow{2}{*}{50\%}                                                \\
                           & Iter. Prun. \cite{han2015learning}             & 92.2\%            &                                                                      \\ \hline\hline
\multirow{2}{*}{ResNet-18} & SLR \textit{(at 20k iterations)} & 89.9\%            & \multirow{2}{*}{11.4\%}                                              \\
                           & Iter. prun. \cite{frankle2018lottery}             & 75.0\%            &                                                                      \\ \hline\hline
\multirow{2}{*}{ResNet-50} & \textbf{SLR}            & 93.6\%            & \multirow{2}{*}{40\%}                                                \\
                           & AMC \cite{He_2018_ECCV}                    & 93.5\%            &                                                                      \\ \hline
\end{tabular}
\end{adjustbox}
\end{table}

Table \ref{table:slr_comp} shows our comparison of SLR with other recent model compression works on CIFAR-10 benchmark. We report the weights remaining after SLR training and the final accuracy. We start training the networks with a learning rate of $0.1$ decreasing the learning rate by a factor of 10 at epochs 80 and 100. On ResNet-18, we compare our result at only 20k iterations. For VGG-16 and ResNet-50, we observe that SLR can achieve up to 60\% pruning with less than 1\% accuracy drop from the baseline model.
\subsection{Evaluation on Object Detection Tasks}\label{object}

\noindent\textbf{Models and Datasets.} We used YOLOv3 and YOLOv3-tiny models \cite{yolov3} on COCO 2014 benchmark. We used and followed the publicly available Ultralytics repository\footnote{\url{https://github.com/ultralytics/yolov3}} for YOLOv3 and its pretrained models. For lane detection experiment, we used the pretrained model from Ultra-Fast-Lane-Detection~\cite{qin2020ultra} on TuSimple lane detection benchmark dataset. \\
\noindent\textbf{Training Settings.} In all experiments we used $\rho = 0.1$. We set SLR parameters as $M = 300$, $r = 0.1$ and $s_0 = 10^{-2}$. We follow the same training settings provided by the repositories we use. Finally, we use the same number of training epochs and sparsity configuration for ADMM and SLR.\\
\noindent\textbf{Testing Settings.} On YOLOv3 models, we calculate the COCO mAP with IoU = 0.50 with image size of 640 for testing. On lane detection experiments, evaluation metric is ``accuracy", which is calculated as $\frac{\sum_{\text {clip}} C_{\text {clip}}}{\sum_{\text {clip}} S_{\text {clip}}}$, where $C_{clip}$ is the number of lane points predicted correctly and $S_{clip}$ is the total number of ground truth in each clip.\\

\noindent\textbf{Evaluation of SLR Performance.}\label{object_results}
Our comparison of SLR and ADMM methods on COCO dataset is shown in Table \ref{table:addmvsslr_coco}. We have compared the two methods under 3 different compression rates for YOLOv3-tiny and tested YOLOv3-SPP pretrained model with a compression rate of 1.98$\times$. We can see that the model pruned with SLR method has higher accuracy after hard-pruning in all cases. At a glance at YOLOv3-tiny results, we observe that the advantage of SLR is higher with an increased compression rate.\newline

\begin{table}[!bth]
\caption{ADMM and SLR training results with YOLOv3 on COCO benchmark.}
\label{table:addmvsslr_coco}
\centering
\begin{adjustbox}{width=0.6\columnwidth}
\begin{tabular}{ccccc}
\textbf{Architecture}                                                                            & \textbf{Epoch}      & \textbf{Method} & \textbf{\begin{tabular}[c]{@{}c@{}}Hardpruning \\ mAP\end{tabular}} & \textbf{\begin{tabular}[c]{@{}c@{}}Compression \\ Rate\end{tabular}} \\ \hline\hline
\multirow{2}{*}{}                                                                    & \multirow{2}{*}{15} & ADMM            & 35.2                                                            & \multirow{2}{*}{1.19$\times$}                                        \\
                                                                                     &                     & SLR             & \textit{36.1}                                                   &                                                                      \\ \cline{2-5} 
\multirow{2}{*}{\begin{tabular}[c]{@{}c@{}}YOLOv3-tiny \\ \textit{(mAP = 37.1)}\end{tabular}} & \multirow{2}{*}{20} & ADMM            & 32.2                                                            & \multirow{2}{*}{2$\times$}                                           \\
                                                                                     &                     & SLR             & \textit{36.0}                                                   &                                                                      \\ \cline{2-5} 
\multirow{2}{*}{}                                                                    & \multirow{2}{*}{25} & ADMM            & 25.3                                                            & \multirow{2}{*}{3.33$\times$}                                        \\
                                                                                     &                     & SLR             & \textit{34.0}                                                     &                                                                      \\ \hline\hline
\multirow{2}{*}{\begin{tabular}[c]{@{}c@{}}YOLOv3-SPP \\ \textit{(mAP = 64.4)}\end{tabular}}  & \multirow{2}{*}{15} & ADMM            & 42.2                                                            & \multirow{2}{*}{2$\times$}                                        \\
                                                                                     &                     & SLR             & \textit{53.2}                                                   &                                                                      \\ \hline
\end{tabular}
\end{adjustbox}
\end{table}

\begin{figure}[H]
  \centering
  \subcaptionbox{YOLOv3. \label{fig:yolo_hardprune}}{\includegraphics[width=2.7in]{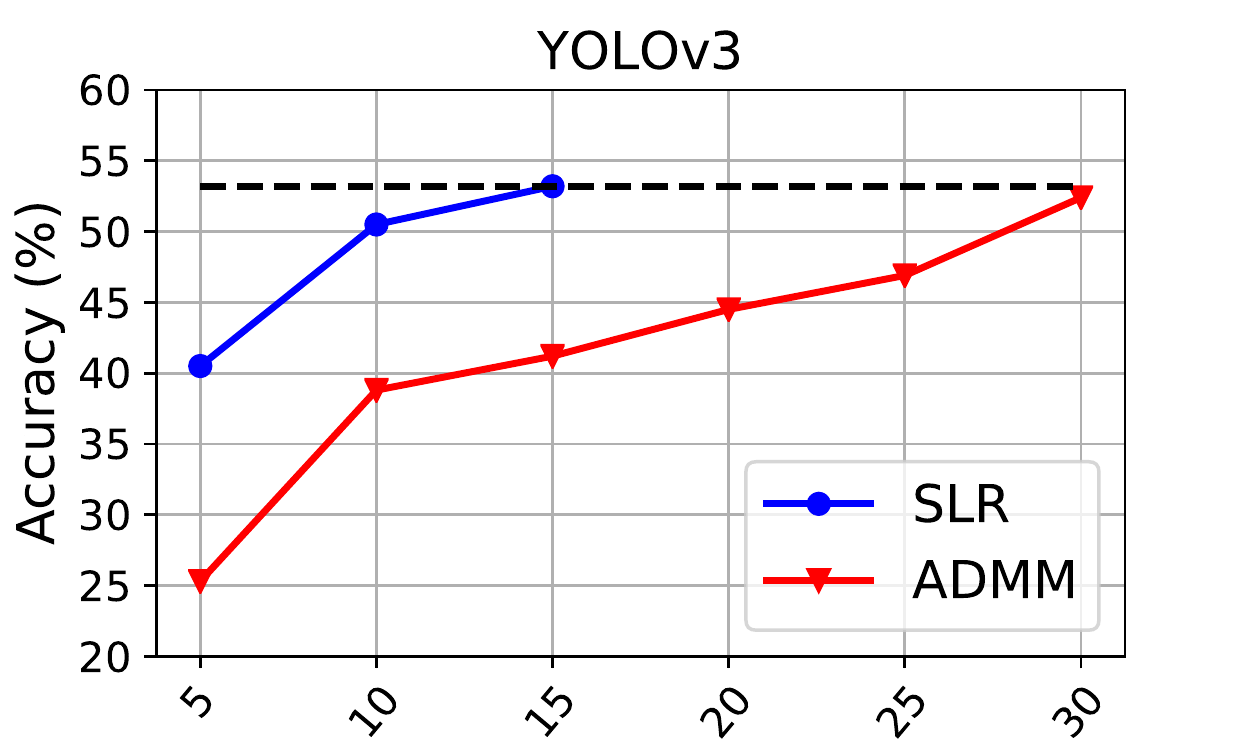}}\hspace{0em}%
  \subcaptionbox{YOLOv3-tiny. \label{fig:yolotiny_hardprune}}{\includegraphics[width=2.7in]{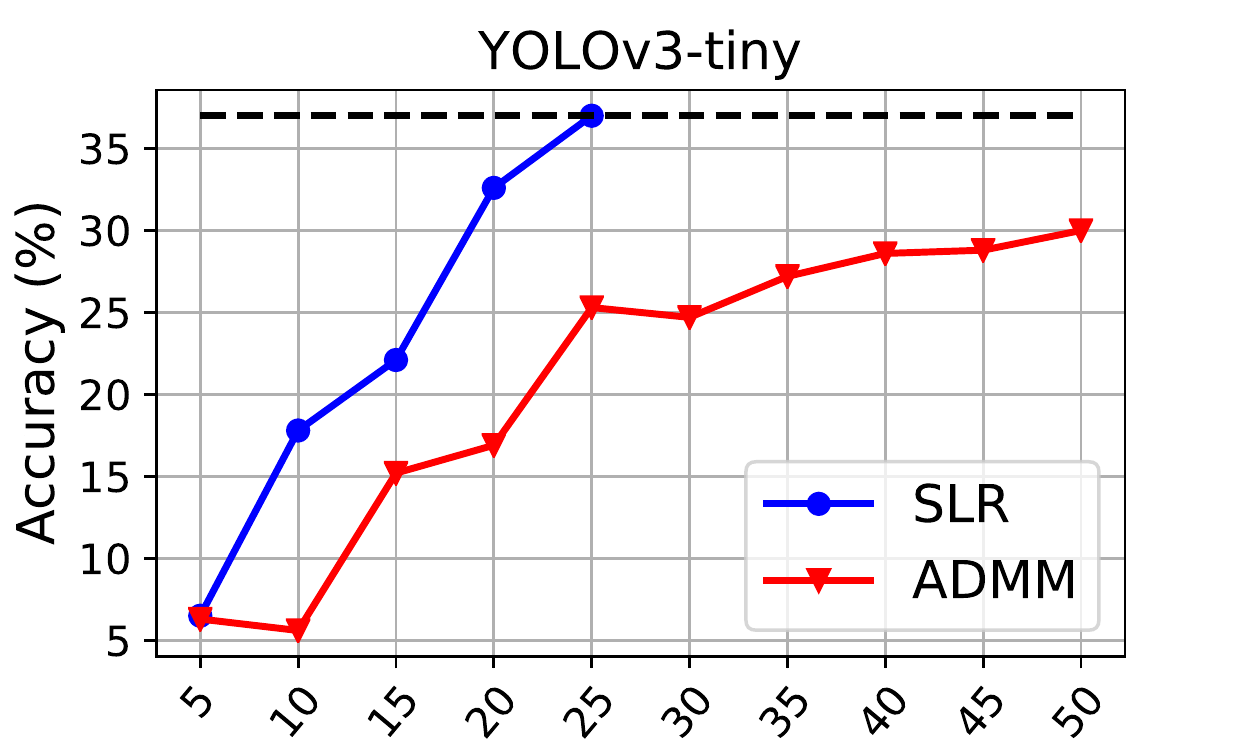}}

  \caption{Hardpruning accuracy of YOLOv3 and YOLOv3-tiny. Accuracy is reported every 5 epochs and training is stopped when methods reach the accuracy threshold.}
\label{fig:hardprune_yolo}
\end{figure}

\noindent In compression rate 3.33$\times$ on YOLOv3-tiny, given a limit of 25 epochs, we can observe that the gap between ADMM and SLR is much higher, which is the due to the faster convergence of SLR as shown in Figure \ref{fig:yolotiny_hardprune}. Similarly, Figure \ref{fig:yolo_hardprune} shows the mAP progress of YOLOv3 during SLR and ADMM training for 50 epochs, pruned with $2\times$ compression. SLR reaches the mAP threshold only at epoch 15.

\begin{figure}[!htb] 
\centering
\includegraphics[width=0.7\columnwidth]{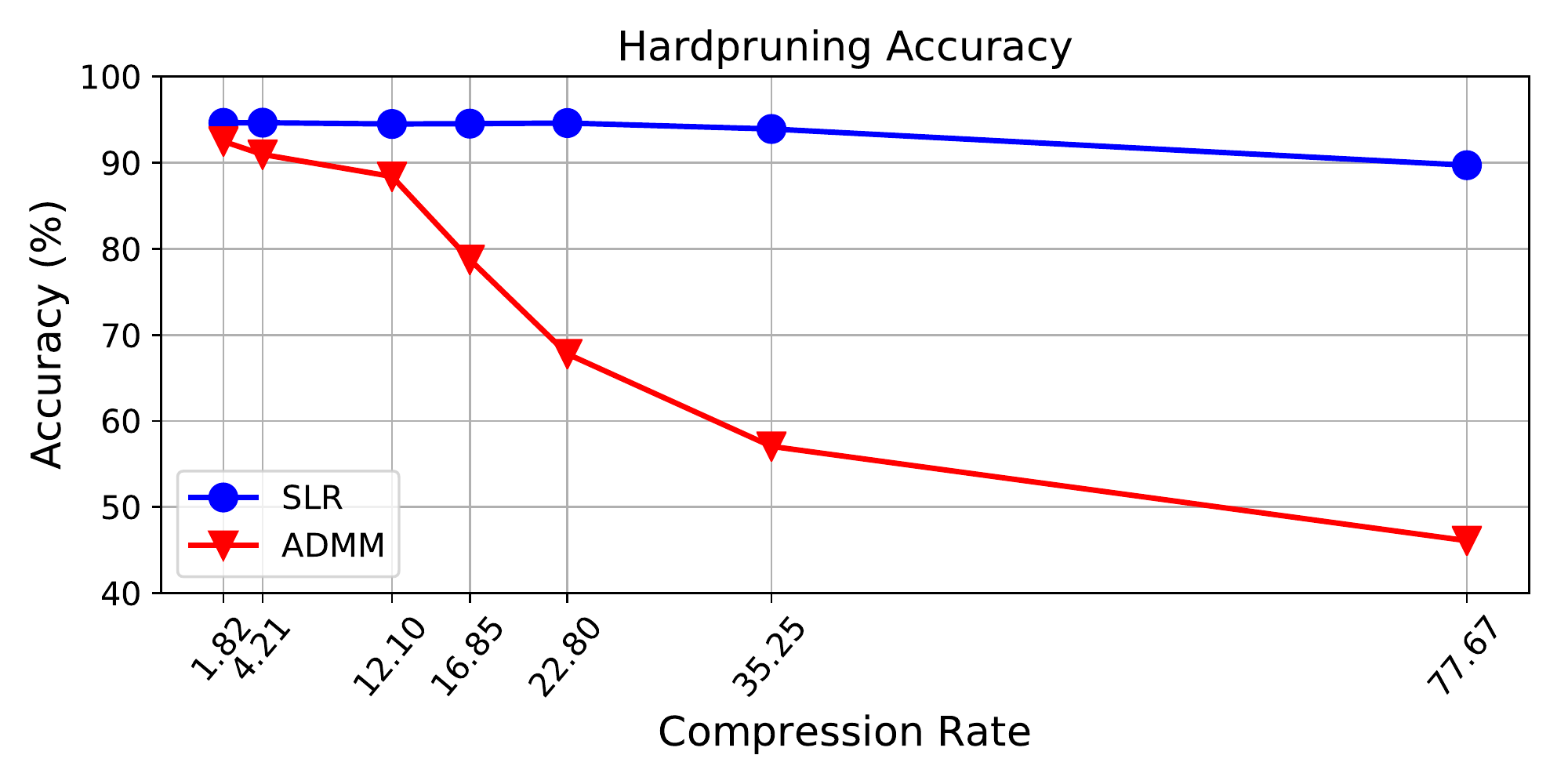}
\caption{Hardpruning accuracy on TuSimple benchmark with ADMM vs. SLR training for several compression rates. SLR has a greater advantage over ADMM as compression rate increases.}
\label{fig:tusimple_comp}
\end{figure}

\noindent Table \ref{table:lane_seg} reports our result for the Lane Detection task on TuSimple lane detection benchmark after 40 epochs of training and 5 epochs of masked-retraining. We conducted experiments under 8 different compression rates. Similarly, Figure \ref{fig:tusimple_comp} illustrates the accuracy gap between ADMM and SLR methods after hard pruning as compression rate increases. 
\begin{table}[htb!]
\caption{SLR pruning results with ResNet-18 on TuSimple benchmark through different compression rates.}
\label{table:lane_seg}
\centering
\begin{adjustbox}{width=0.7\columnwidth,center} 
\begin{tabular}{ccccc}
\hline
\multirow{2}{*}{\textbf{\begin{tabular}[c]{@{}c@{}}Compression\end{tabular}}} & \multicolumn{2}{c}{\textbf{\begin{tabular}[c]{@{}c@{}}Hardpruning Accuracy (\%)\end{tabular}}} & \multicolumn{2}{c}{\textbf{\begin{tabular}[c]{@{}c@{}}Retraining Accuracy (\%)\end{tabular}}} \\ \cline{2-5} 
                                                                                      & ADMM                             & \multicolumn{1}{c|}{SLR}                                      & ADMM                                            & SLR                                            \\ \hline
$1.82\times$                                                                          & \textbf{92.49}                   & \multicolumn{1}{c|}{\textit{\textbf{94.64}}}                  & 94.28                                           & 94.63                                          \\ \cline{2-5} 
$2.54\times$                                                                          & \textbf{92.25}                   & \multicolumn{1}{c|}{\textit{\textbf{94.56}}}                  & 94.04                                           & 94.93                                          \\ \cline{2-5} 
$4.21\times$                                                                          & \textbf{90.97}                   & \multicolumn{1}{c|}{\textit{\textbf{94.66}}}                  & 94.18                                           & 94.68                                          \\ \cline{2-5} 
$12.10\times$                                                                         & \textbf{88.41}                   & \multicolumn{1}{c|}{\textit{\textbf{94.51}}}                  & 94.45                                           & 94.7                                           \\ \cline{2-5} 
$16.85\times$                                                                         & \textbf{78.75}                   & \multicolumn{1}{c|}{\textit{\textbf{94.55}}}                  & 94.23                                           & 94.65                                          \\ \cline{2-5} 
$22.80\times$                                                                         & \textbf{67.79}                   & \multicolumn{1}{c|}{\textit{\textbf{94.62}}}                  & 94.08                                           & 94.55                                          \\ \cline{2-5} 
$35.25\times$                                                                         & \textbf{57.05}                   & \multicolumn{1}{c|}{\textit{\textbf{93.93}}}                  & 93.63                                           & 94.34                                          \\ \cline{2-5} 
$77.67\times$                                                                         & \textbf{46.09}                   & \multicolumn{1}{c|}{\textit{\textbf{89.72}}}                  & 88.33                                           & 90.18                                          \\ \hline
\end{tabular}

\end{adjustbox}
\end{table}

\noindent From Figure \ref{fig:tusimple_comp}, our observation is that for a small compression rate such as $1.82\times$, SLR has little advantage over ADMM in terms of hardpruning accuracy. However, as the compression rate increases, SLR starts to perform better than ADMM. For example, SLR survives $77.67\times$ compression with slight accuracy degration and results in 89.72\% accuracy after hardpruning while ADMM accuracy drops to 46.09\%. This demonstrates that our SLR-based training method has a greater advantage over ADMM especially in higher compression rates, as it achieves compression with less accuracy loss and reduces the time required to retrain after hard-pruning.

\begin{figure}[tbh!]
  \centering
  \subcaptionbox{Weights before after pruning with SLR (middle) and ADMM (right) under the same compression rate (77.6$\times$). \label{fig3:tusimple_a}}{\includegraphics[width=6in]{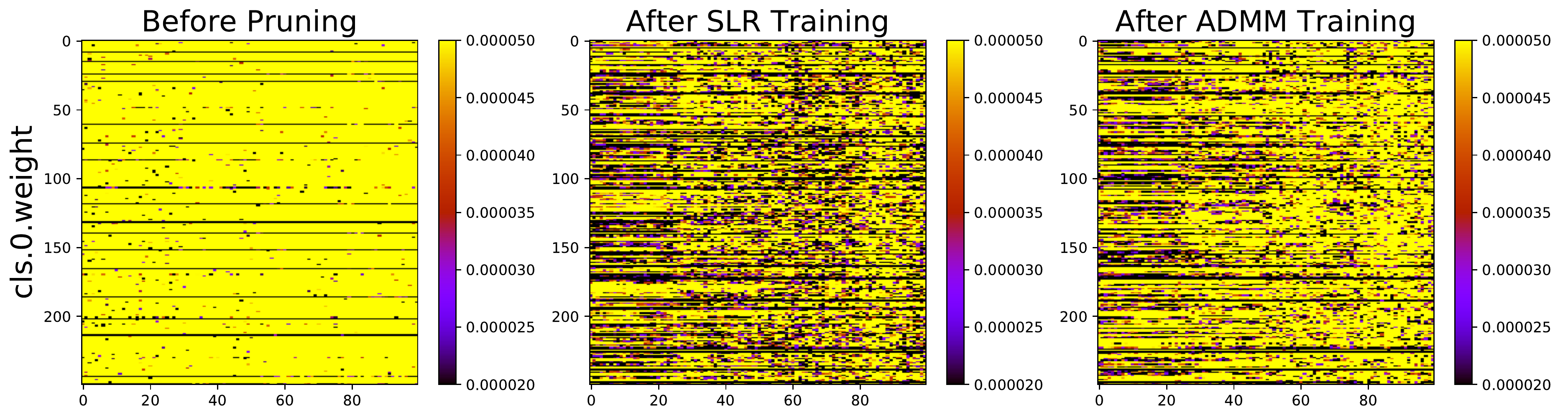}}\hspace{0em}%
  \subcaptionbox{Weights before and after pruning with SLR (middle) and ADMM (right) under the same accuracy (89.0\%). \label{fig3:tusimple_b}}{\includegraphics[width=6in]{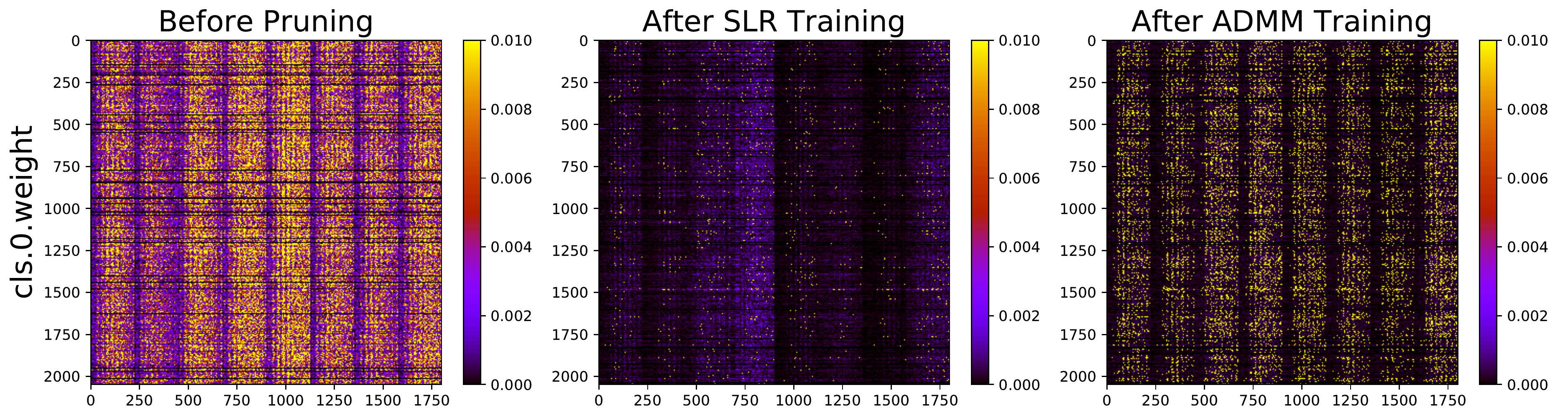}}
  \caption{Heatmap of ResNet-18 weights on TuSimple benchmark before and after pruned with SLR and ADMM. Weights are more zeroed out with SLR compared to ADMM. }
\label{fig:tusimple_heat}
\end{figure}

\noindent Finally, in Figure \ref{fig:tusimple_heat}, we show the difference between weights of one layer before and after pruning with SLR and ADMM. In Figure \ref{fig3:tusimple_a}, we initially show the initial (non-pruned) weights and then show the sparsity of weights under the same compression rate (77$\times$) with SLR and ADMM. Initially, the layer has low sparsity. After training with SLR and ADMM, we can see an increased number of zeroed-weights. SLR moves towards the desired sparsity level faster than ADMM. In Figure \ref{fig3:tusimple_b}, we compare the sparsity of weights under the same accuracy (89.0\%). It can be observed that SLR significantly reduced the number of non-zero weights and ADMM has more non-zero weights remaining compared with SLR.

\subsection{Ablation Studies}\label{ablation}

\begin{figure}[htb!]
  \centering
  \subcaptionbox{$s_0$ parameter. \label{fig3:finetuning_s}}{\includegraphics[width=2in]{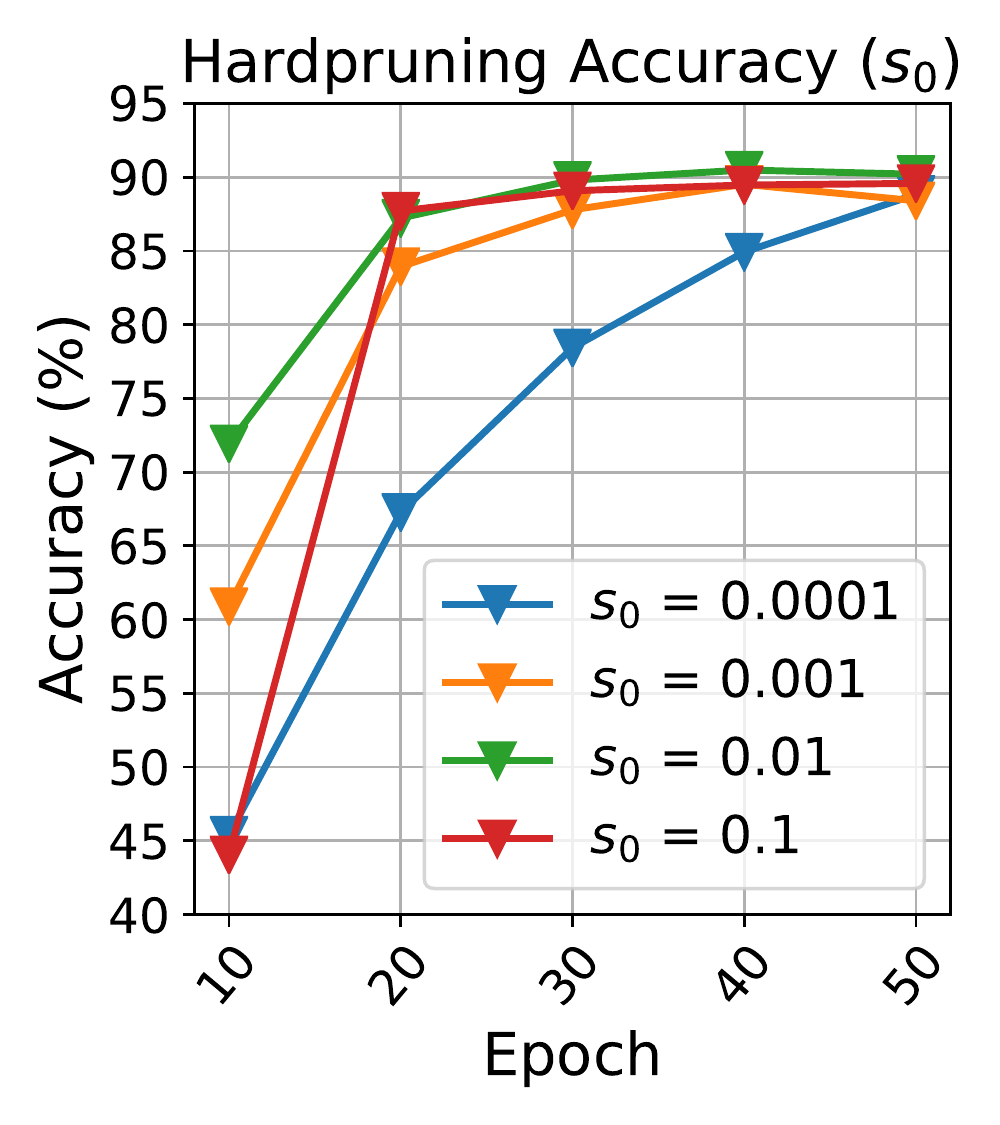}}\hspace{0em}%
  \subcaptionbox{$M$ parameter. \label{fig3:finetuning_m}}{\includegraphics[width=2in]{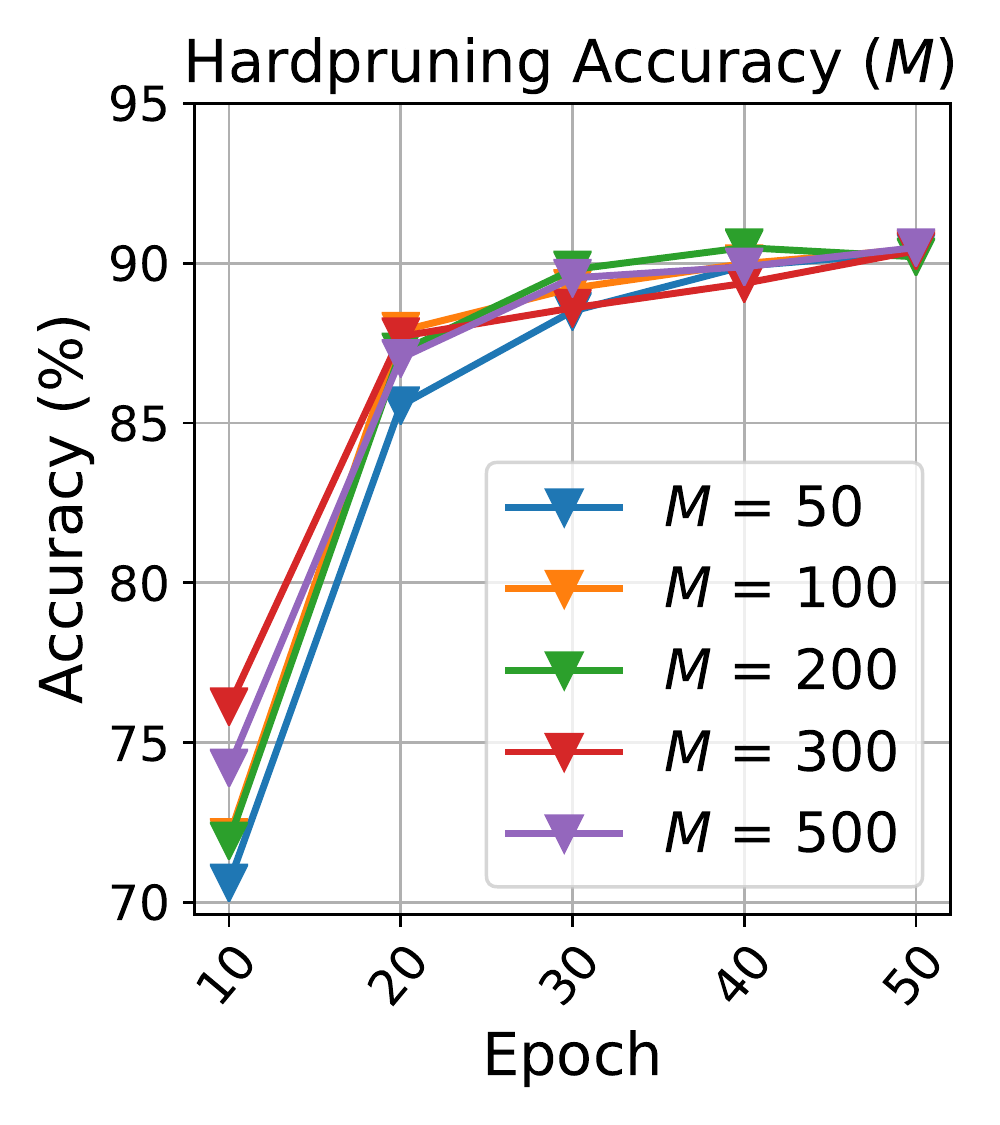}}\hspace{0em}%
  \subcaptionbox{$r$ parameter.\label{fig3:finetuning_r}}{\includegraphics[width=2in]{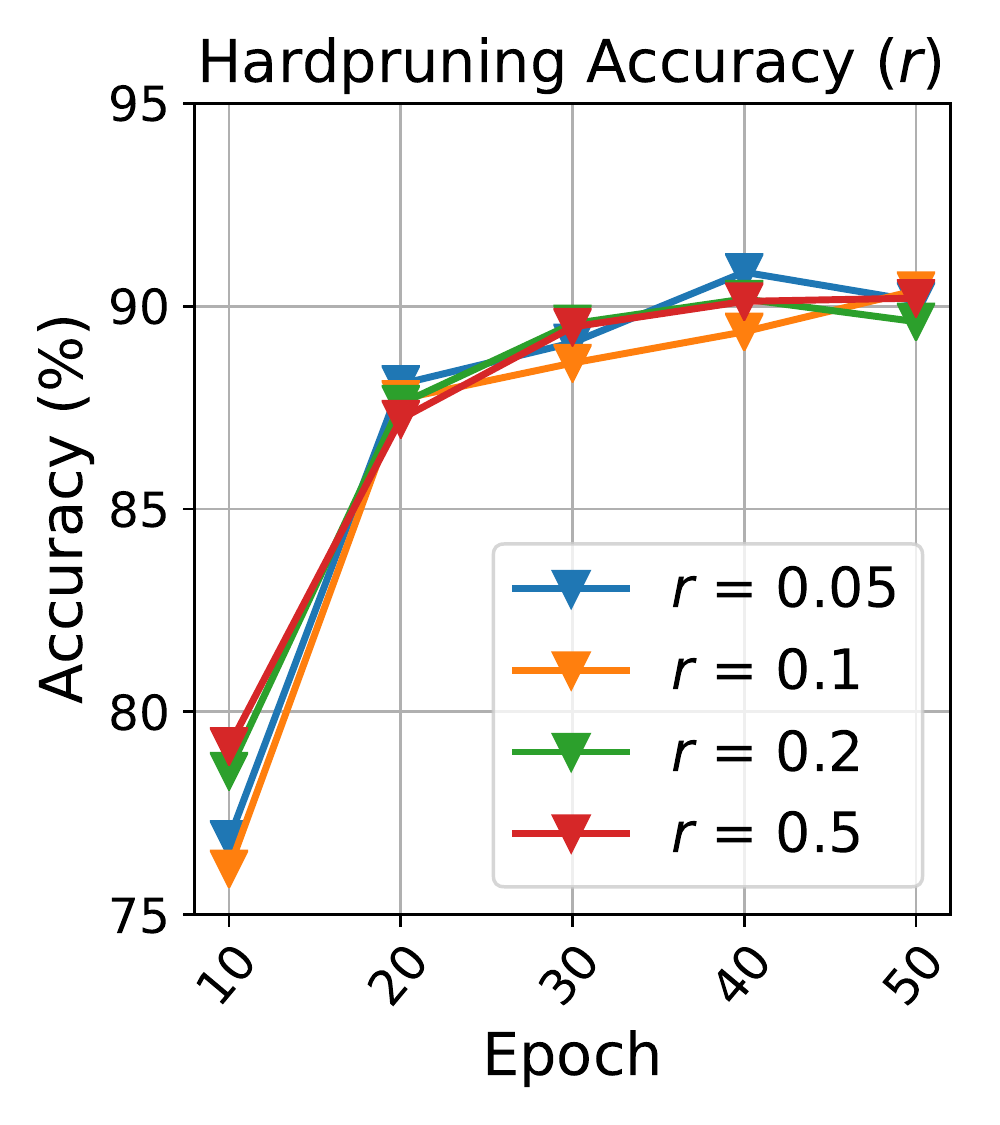}}

  \caption{Hardpruning accuracy of ResNet-18 on CIFAR-10 during SLR training with respect to different values of $s_0$, $M$ and $r$.}
\label{fig:finetuning_figs}
\end{figure}

We conducted several experiments to observe SLR behavior with respect to SLR parameters $\rho, s_0, r$ and $M$ on ResNet-18 model (93.33\% accuracy) and CIFAR-10. We pruned the model through SLR training for 50 epochs with a compression rate of $8.71\times$ and observed the hardpruning accuracy every 10 epochs. Figure \ref{fig:finetuning_figs} shows the accuracy of the model through SLR training based on the different values of $s_0$, $M$ and $r$. Based on the hardpruning accuracy throughout training, it can be seen that, even though the parameters do not have a great impact on the end result, choice of $s_0$ can impact the convergence of the model. From Figure \ref{fig3:finetuning_s}, we can state that $s_0 = 10^{-2}$ provides higher starting accuracy and converges quickly. Figure \ref{fig3:finetuning_m} and Figure  \ref{fig3:finetuning_r} demonstrate the impact of $M$ and $r$ on the hardpruning accuracy respectively.\\ 

\begin{figure}[!tbh]
\centering
\includegraphics[width=0.7\columnwidth]{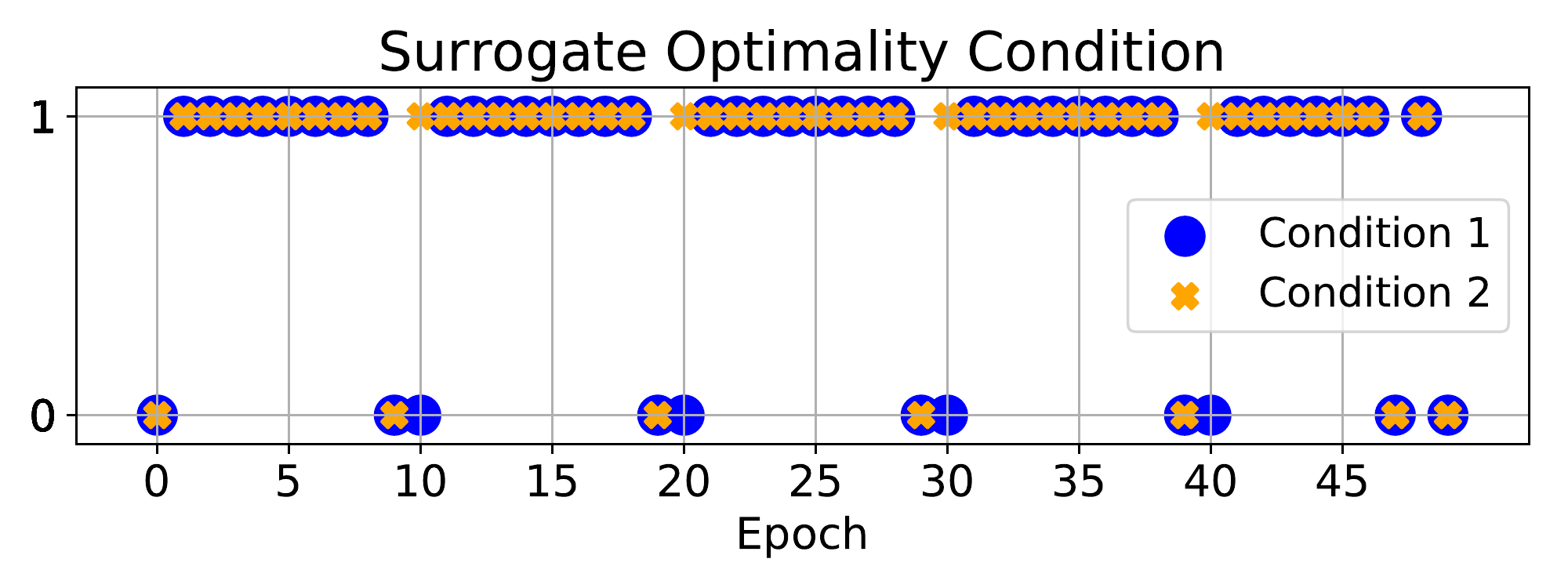}
\caption{Surrogate optimality condition satisfaction graph during the SLR training of ResNet-18 on CIFAR-10 for 50 epochs (1: satisfied, 0: not satisfied). The condition is satisfied periodically.}
\label{fig:opt_condition_50epochs}
\end{figure}

\noindent Figure \ref{fig:opt_condition_50epochs} demonstrates that there exists iteration $\kappa$ (as required in the Theorem) so that the surrogate optimality condition, the high-level convergence criterion of the SLR method, is satisfied during training with $s_0 = 10^{-2}$, $\rho=0.1$ thereby signifying that ''good" multiplier-updating directions are always found. For example, after the conditions are violated at epoch 9, there exits $\kappa = 10$ so that at iteration 11, after $\kappa = 10,$ the surrogate conditions are satisfied again.  \newline

\section{Conclusions}
In this paper, we presented the DNN weight-pruning problem as a non-convex optimization problem by adopting the cardinality function to induce weight sparsity. By using the  SLR method, the relaxed weight-pruning problem is decomposed into subproblems, which are then efficiently coordinated by updating Lagrangian multipliers, resulting in fast convergence. We conducted weight-pruning experiments on image classification and object detection tasks to compare our SLR method against ADMM. We observed that SLR has a significant advantage over ADMM under high compression rates and achieves higher accuracy during weight pruning. SLR reduces the accuracy loss caused by the hard-pruning and so shortens the retraining process. With the effective optimization capabilities through coordination with clear advantages shown from several examples, the SLR method has a strong potential for more general DNN-training applications. 

{\small
\bibliographystyle{unsrt}
\bibliography{bibliography}
}

\end{document}